\documentclass[12pt]{article}
\usepackage[utf8]{inputenc}
\usepackage[T1]{fontenc}
\usepackage{authblk}
\usepackage{graphicx}
\graphicspath{{figures/}}
\usepackage{natbib}
\usepackage{hyperref}
\usepackage{amssymb}
\usepackage{amsmath}
\usepackage{booktabs}
\usepackage{subcaption}
\usepackage{geometry}
\usepackage{setspace}
\usepackage[colorinlistoftodos,prependcaption]{todonotes}
\usepackage{parskip}
\usepackage{algorithm}
\usepackage{algpseudocode}
\usepackage{enumerate}
\usepackage{ragged2e}
\usepackage{comment}
\usepackage{tablefootnote}
\usepackage{url}

\title{Non-myopic Matching and Rebalancing in Large-Scale On-Demand Ride-Pooling Systems Using Simulation-Informed Reinforcement Learning}

\author[a]{Farnoosh Namdarpour}
\author[a]{Joseph Y. J. Chow}

\affil[a]{C2SMARTER University Transportation Center, New York University Tandon School of Engineering, Brooklyn, NY, USA}

\date{October 2025}

\begin{document}

\maketitle

\newpage

\begin{abstract} 
Ride-pooling, also known as ride-sharing, shared ride-hailing, or microtransit, is a service wherein passengers share rides. This service can reduce costs for both passengers and operators and reduce congestion and environmental impacts. A key limitation, however, is its myopic decision-making, which overlooks long-term effects of dispatch decisions. To address this, we propose a simulation-informed reinforcement learning (RL) approach. While RL has been widely studied in the context of ride-hailing systems, its application in ride-pooling systems has been less explored. In this study, we extend the learning and planning framework of \citet{xu2018large} from ride-hailing to ride-pooling by embedding a ride-pooling simulation within the learning mechanism to enable non-myopic decision-making. In addition, we propose a complementary policy for rebalancing idle vehicles. By employing n-step temporal difference learning on simulated experiences, we derive spatiotemporal state values and subsequently evaluate the effectiveness of the non-myopic policy using NYC taxi request data. Results demonstrate that the non-myopic policy for matching can increase the service rate by up to 8.4\% versus a myopic policy while reducing both in-vehicle and wait times for passengers. Furthermore, the proposed non-myopic policy can decrease fleet size by over 25\% compared to a myopic policy, while maintaining the same level of performance, thereby offering significant cost savings for operators. Incorporating rebalancing operations into the proposed framework cuts wait time by up to 27.3\%, in-vehicle time by 12.5\%, and raises service rate by 15.1\% compared to using the framework for matching decisions alone at the cost of increased vehicle minutes traveled per passenger.
\end{abstract}

\clearpage

\section{Introduction} \label{sec:Introduction}

Ride-pooling services, such as UberX Share\citep{Uber}, Via Microtransit\citep{Via},  MOIA\citep{MOIA}, and GrabShare \citep{Grab}, have expanded their operations over the years. Ride-pooling or ride-sharing refers to a system where passengers with different ride requests share one vehicle. Compared to ride-hailing where only one passenger is onboard the vehicle with the driver, ride-pooling can offer increased efficiency for service providers and more affordable rides for customers, while reducing traffic congestion and environmental impacts. However, dispatching decisions, i.e. how vehicles are assigned to requests and repositioned across space, is central to system performance.  Operators value higher ridership and reduced vehicle distance traveled, while riders prioritize shorter waiting and in-vehicle times. Additionally, in a ride-pooling system, considering passengers already onboard the vehicle adds another layer of complexity when finding efficient matches for future requests. The system's efficiency depends on an optimization algorithm capable of balancing the priorities of both riders and operators. 

In a dynamic ride-pooling system, ride requests are submitted over time while the system attempts to find a vehicle for each request once submitted. Additionally, idle vehicles across the service area are generally relocated to better match supply with demand.
The vehicle-request matching and the vehicle rebalancing (also known as repositioning or redistribution) problems are both sequential decision problems. Decisions made at a certain time can impact the system in the future, highlighting the importance of considering the long-term impact of decisions. Among the non-myopic optimization methods for sequential decision making, reinforcement learning (RL) has shown promising results \citep{qin2022reinforcement}. In RL, an agent interacts with the environment by receiving feedback signals in the form of reward, which indicates how well the agent is performing. The agent's goal is to maximize the cumulative reward. 

While RL has been widely applied to ride-hailing systems, it is less explored in ride-pooling systems due to the more complex nature of the problem. Most existing RL-based papers on ride-pooling treat each vehicle as an individual agent, applying the trained model independently to each one \citep{qin2022reinforcement, al2019deeppool}. However, Didi's statistics have shown that centralized fleet dispatch, where the platform assigns vehicles to riders instead of vehicles being the decision-makers, can significantly improve system efficiency \citep{xu2018large}. \citet{xu2018large} proposed a learning and planning approach for dispatching vehicles in a ride-hailing system to optimize long-term global efficiency, which is applicable to large-scale platforms. Their proposed method was deployed in the production system of Didi Chuxing. 

We build upon their work and extend their methodology in a novel simulation-informed manner to ride-pooling systems. This is one of the first studies to propose a non-myopic RL approach for real-time dispatch in large-scale ride-pooling systems with a central dispatch unit. We propose an offline approach to learn the spatiotemporal patterns of supply and demand from episodes of experience generated in a simulated ride-pooling environment with historical demand data and propose an online non-myopic policy to use the learned value functions from this simulation-informed process for making real-time dispatch decisions. The sample-based learning approach provides an efficient and powerful method for leveraging simulation to generate samples and learn value functions from the simulated experiences. The proposed methodology is evaluated using the NYC yellow taxi request data \citep{NYCTaxiData} and a ride-pooling simulator (NOMAD-RPS) developed by \citet{namdarpour2024non}. The results show significant improvements over baseline algorithms in terms of both operator metrics (service rate) and passenger metrics (wait time and in-vehicle time), which remain consistent across different tested fleet sizes. Our contributions are summarized below.

\begin{itemize}
    \item Propose an offline policy evaluation method using n-step temporal difference (TD) learning \citep{sutton2018reinforcement} to learn spatiotemporal value functions from episodes of experience generated by historical demand data that are simulated within a ride-pooling simulator with a hyperparameterized fleet size.
    \item Propose online policies that use the learned value functions and immediate rewards to make real-time dispatch decisions in large-scale ride-pooling systems, one policy dedicated to optimizing vehicle-rider matching decisions and another to guiding vehicle rebalancing decisions. 
    \item Evaluate the proposed framework using a large-scale real-world taxi dataset and simulator.
\end{itemize}

The remainder of the paper is organized as follows. Section \ref{sec:LiteratureReview} reviews the literature on dispatching vehicles in ride-pooling systems. Section \ref{sec:Methodology} introduces the proposed methodology for non-myopic dispatch in large-scale ride-pooling systems. Experimental results using NYC taxi data are presented in Section \ref{sec:ComputationalExperiments}, and Section \ref{sec:Conclusion} concludes the paper. 

\section{Literature review} 
\label{sec:LiteratureReview}

Traditional approaches for making dispatch decisions in ride-pooling systems often model the problem as an optimization problem \citep{ho2018survey}. However, these approaches are generally not scalable for large-scale on-demand operations \citep{shah2020neural}. Another group of studies uses greedy algorithms to find the best match, such as finding the nearest vehicle to serve a request. Although these approaches typically scale well, they are myopic and ignore the impact of decisions on the future horizon (\citep{alonso2017demand, santi2014quantifying, jung2016dynamic}).

More recent studies use the Markov decision process framework to find non-myopic solutions to the problem. However, the literature on non-myopic approaches is mostly limited to ride-hailing systems, as the state and action spaces in ride-pooling systems can grow exponentially, making the problem more complex. Among the studies focused on ride-pooling, some used cost function approximation (CFA) policies that involve tuning parameters and searching among a family of functions using an objective
function to find the best policy \citep{powell2019unified}. 

\citet{hyytia2012non} proposed a non-myopic CFA policy by modeling the dynamic pickup and delivery problem as a multi-server queue system and approximating the system over an infinite horizon. This method has been used in multiple studies, such as \citet{sayarshad2015scalable,ma2019dynamic,namdarpour2024non}. \citet{ma2019dynamic} integrated a ride-sharing system with public transit using queueing-theoretic vehicle dispatch and idle vehicle relocation algorithms. \citet{namdarpour2024non} proposed a non-myopic CFA policy for operating a ride-pooling system with synchronized transfers, allowing passengers to transfer between vehicles within the system. 
While CFA policies are suitable for large-scale problems, their performance may be compromised compared to other Reinforcement Learning (RL) methods.

Among other studies using approximate dynamic programming (ADP) and RL that incorporate value functions, one group treats each vehicle in the system as an individual agent while there is no coordination among them \citep{gueriau2018samod, jindal2018optimizing, al2019deeppool, haliem2021adapool}. Another group of studies considers centralized fleet dispatch\citep{yu2019integrated, shah2020neural, li2022value}, where vehicles are assigned to requests by a central agent. 

In the first group, using a decentralized Q-learning multi-agent approach, \citet{gueriau2018samod} proposed SAMoD, Shared Autonomous Mobility-on-Demand, which integrates rebalancing and request assignment. Using RL and deep neural networks, \citet{al2019deeppool} proposed a model-free framework called DeepPool to dispatch vehicles where each vehicle trains its own deep Q-network independently, resulting in substantial reductions in complexity through decentralized learning. \citet{haliem2021adapool} proposed AdaPool, an adaptive matching and dispatching framework using deep RL, which deals with highly dynamic environments by using an online Dirichlet change point detection and adapting Deep Q-learning to develop optimal policies tailored to various environment models. \citet{singh2021distributed} and \citet{wang2023optimization} used deep RL for solving the vehicle dispatching problem with unsynchronized transfers where passengers are allowed to transfer between vehicles in a ride-pooling service.

Among the studies in the second group, \citet{yu2019integrated} used ADP to model the ride-pooling problem where no more than two passenger groups share rides at the same time. They developed a heuristic decomposition scheme to enhance computational efficiency. \citet{shah2020neural} also proposed an ADP method, where matching was conducted using a value function learned and approximated by neural networks to estimate the future value of each matching decision. Following \citet{shah2020neural}, \citet{li2022value} developed an offline policy evaluation-based method to learn value functions more efficiently using neural networks and adopted an online learning procedure to update the learned values in real time and use the learned values for batch assignment. \citet{tang2021value} used centralized value functions to make both dispatching and repositioning decisions in a ride-hailing system and proposed a value ensemble approach that combines online learning with offline training. \citet{liu2024joint} introduced a directional state capturing travel direction and used deep reinforcement learning to learn value functions in a ride-pooling setting through offline training, which were then applied for batch order assignment and idle vehicle rebalancing, with rebalancing restricted to vehicles' current or neighboring grids.

While the first group of studies can benefit from computational efficiency through decentralized learning, the second group can achieve greater improvements in system efficiency \citep{xu2018large}. However, very few studies have formulated the system with a central dispatch unit as an RL problem for a ride-pooling system; these approaches often suffer from long training times and lack scalability for large-scale systems, or train on historical dispatch data that does not necessarily represent ideal conditions. 

While not directly related to ride-pooling, applications of physics-informed reinforcement learning have gained traction in recent years \citep{shi2021physics, han2022physics} in which RL is combined with model-based approaches to leverage the structure of the problem captured by the model. In this study, we build upon the work of \citet{xu2018large}, which uses RL for dispatch in ride-hailing systems, but extend it to ride-pooling systems using simulation models to help inform on the structure of the decision for the learning mechanism, i.e. simulation-informed. The spatiotemporal value functions are learned from episodes generated using historical demand data simulated in a ride-pooling environment with a hyperparameterized fleet size. The simulator’s fleet size is set relatively large to avoid request rejections, allowing the model to capture spatiotemporal demand–supply patterns in a more idealized setting. In addition to the policy for vehicle–rider matching, we also propose a complementary policy for rebalancing idle vehicles. The proposed non-myopic approach is an effective method that optimizes the long-term system efficiency and is applicable to large-scale real-time systems. 
\section{Methodology} \label{sec:Methodology}

\subsection{Problem description} \label{subsec:ProblemDescription}

An on-demand ride-pooling service is considered to operate in a specified service region. Ride requests are submitted throughout the service operating hours while a centralized dispatch unit matches vehicles with ride requests in real-time with the goal of maximizing one or more performance metrics, such as service rate. Vehicles have flexible routes, allowing passengers to share rides, meaning that other passengers can be picked up or dropped off while a passenger is onboard. Each ride request $r_i$ includes submission time $t_{sub,i}$, origin location, and destination location. It is assumed that each request corresponds to a single passenger who is ready for immediate pickup upon request submission. In the first part, repositioning of idle vehicles is not considered, whereas it is incorporated in the second part, as detailed in Section \ref{subsec: OnlinePlanningRebalancing}.

The road network is represented as a directed graph $G=(N,E)$ with $N$ nodes and $E$ edges. Each edge has a weight corresponding to its travel cost, typically represented by travel time. Virtual stops are defined as a set of nodes where passengers can either be picked up or dropped off. Upon request submission, the origin and destination locations are assigned to nearby virtual stops, denoted as $o_i$ and $d_i$, respectively. 

Vehicles can initially be located randomly at any network node or at predetermined hub locations. The system updates at fixed time intervals $\Delta t$, during which submitted trip requests are forwarded to a centralized system for vehicle assignment. 
To find a feasible match between a vehicle and a rider, the following constraints should be met:
    \begin{itemize}
        \item Passenger's wait time (time difference between request submission time and pickup time) should not be longer than a maximum wait time ($w_{max}$).
	  \item Since other passengers may be picked up or dropped off along the way, a maximum detour delay is defined for each passenger meaning their drop-off time should be earlier than their latest allowed drop-off time. ($t_{latest\: dropoff, i}$).
	  \item Vehicle capacity should not be exceeded at any time.
    \end{itemize}

If the system cannot find a vehicle meeting all the feasibility criteria, the request status changes to \textit{pending}. The system will continue to retry finding a match for a pending request in subsequent time steps until a match is found or the maximum wait time for the passenger has been exceeded. In the latter case, the request will be \textit{rejected}. 

\subsection{Problem formulation} \label{subsec:ProblemFormulation}

Vehicle dispatching in a ride-pooling system is a sequential decision problem that can be modeled as a Markov Decision Process (MDP). In an MDP, an agent interacts with the environment and takes an action $a_t$ at state $s_t$ according to a policy $\pi$. The agent's goal is to maximize the discounted accumulated rewards from time $t$, which is defined as the expected discounted return $G=\sum_{k=0}^{\infty}\gamma^k r_{t+1+k}$ where $\gamma$ is the discount factor and $r_{t+1+k}$ is the immediate reward at time $t+1+k$. The parameter $\gamma$ typically ranges between 0 and 1, with values closer to 0 indicating a myopic agent and values closer to 1 indicating a far-sighted agent. The main components of the MDP are defined below.

\textbf{Agent}: Although from a global perspective, there is only a single centralized agent in the system making dispatch decisions, similar to \citet{xu2018large}, we define the system from a local viewpoint to simplify the definition of model components. In a local view, each vehicle is considered an agent, though we do not distinguish between individual vehicles. The environment contains the state information of all agents in the platform. Agent and vehicle are used interchangeably throughout the rest of the paper.

\textbf{State}: The service operating time is divided into several time periods, e.g. five-minute intervals, and the service area is divided into multiple zones. The state of each agent is defined by the spatiotemporal status of the vehicle, represented by the time index $t$ and the zone index $z$; $s=(t,z) \in S$. The size of the state space is determined by the product of the number of time periods and the number of regions; $|S|=|T| \times |Z|$.

\textbf{Action}: Two actions are defined for each agent. One is to serve a new request and update the vehicle's scheduled route by adding the pickup and dropoff location of the new request. For vehicle–rider matching, the request corresponds to a rider’s trip request, whereas for vehicle repositioning, the request represents the target location to which the vehicle should be relocated. The other action for the agent is to continue its previous schedule without any changes. 

\textbf{Reward}: The reward for an agent at time index $t$ is defined as the number of new requests that have been assigned to the agent in that time index. Therefore, the reward can take zero or positive integer values. By this definition, vehicles learn over time to be in the zones at times they are needed. This definition reflects the optimization goal of the system, which is maximizing the number of served requests while minimizing passengers' travel times.

\textbf{State transition}: According to the time passed in the system, vehicles' locations may change, some passengers are picked up, some are dropped off, and the states change accordingly. 

\subsection{Proposed algorithm} \label{subsec:ProposedAlgorithm}
\begin{figure}
    \centering
    \includegraphics[scale=0.5]{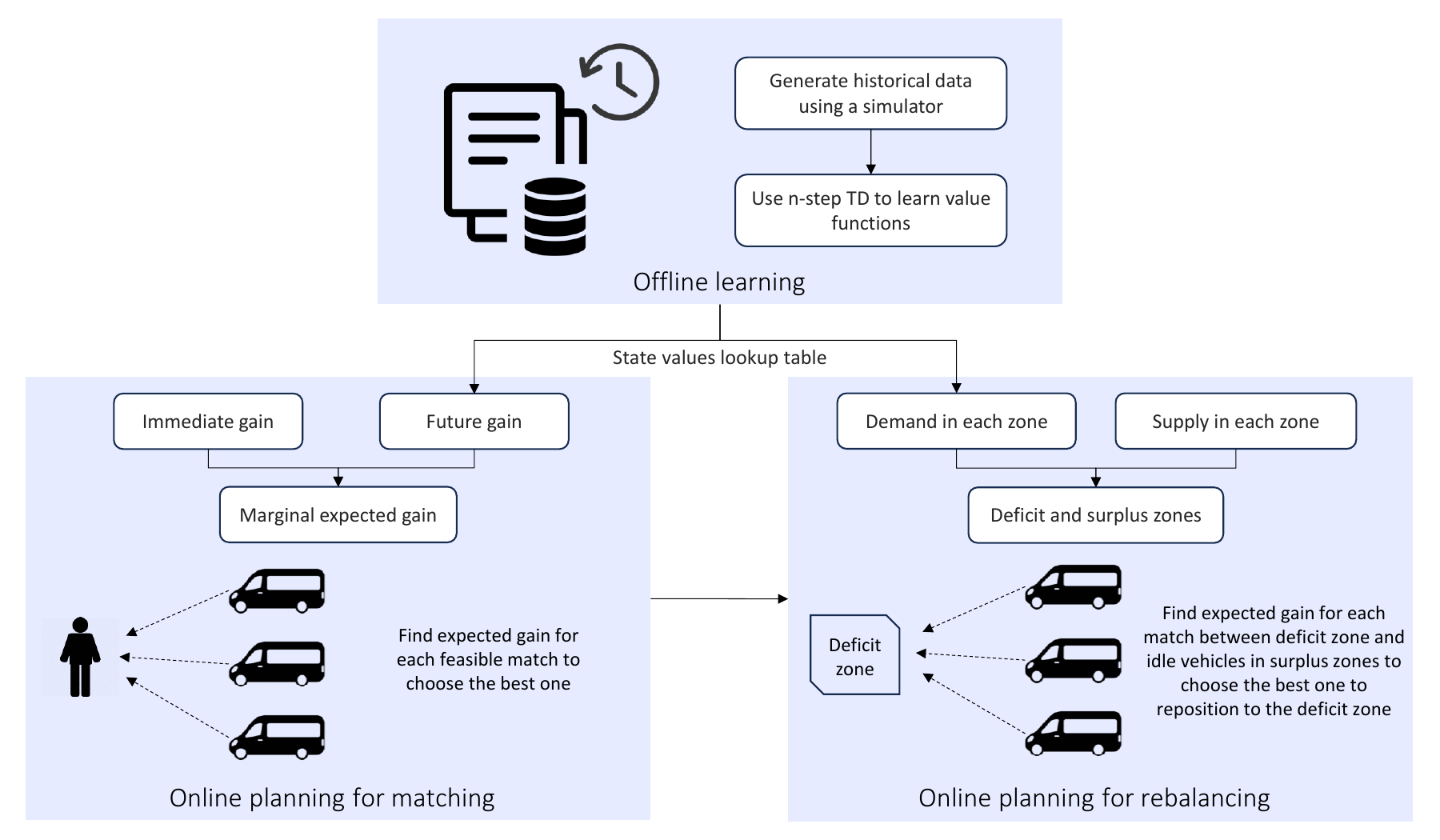}
    \caption{Proposed framework.}
    \label{fig:proposedFramework}
\end{figure}

The proposed framework is a learning and planning approach with three main components shown in Figure \ref{fig:proposedFramework}: offline learning, online planning for matching, and online planning for rebalancing. In offline learning, the state value functions are learned using episodes of experience generated by a ride-pooling service simulator. The online planning components use the learned state values to optimize the system over a long-term horizon in real time. Each component is explained in more details below.

\subsubsection{Offline simulation-informed learning} \label{subsec: OfflineLearning}
A sample-based approach is employed for offline learning, where historical demand data is fed into a ride-pooling simulator with a fleet size hyperparameter to generate simulated experiences. The simulator fleet size is set to a large value to ensure no request rejections occur in the system, thereby capturing all spatiotemporal demand-supply patterns. This can only be effectively achieved in a simulated environment, where control over variables allows for a thorough exploration of different possible scenarios, including the one with a large fleet size that eliminates rejections. A model-free RL method is then applied to these simulation-informed samples to learn value functions. The demand information, including request submission times and pickup and dropoff locations, serves as the input data for the simulator. This information is derived from available field data. All the supply-side information is handled by the simulator. The simulator follows a fixed policy, e.g. a myopic policy, for assigning vehicles to passengers, with a sufficiently large fleet size to ensure that all requests are served, as explained above. For example, the simulator developed by  \citet{namdarpour2024non} has the option of using the myopic policy shown in Eq. \ref{eq:myopicPolicy}.

\begin{equation}
    \label{eq:myopicPolicy}
    c(v,\xi)= \theta \cdot T(v,\xi)  + (1-\theta) \left( \sum_n J_{in-vehicle,n} (v,\xi) + \alpha \sum_n J_{wait,n} (v,\xi) \right )
\end{equation}

where $c(v,\xi)$ is the cost value of the routing/dispatching decision (vehicle $v$, route $\xi$), $T(v,\xi)$ represents the operator cost, which is vehicle $v$’s travel time given route $\xi$. Parameter $\theta$ is the degree of operator cost versus user cost. User cost is represented by passengers’ perceived travel time, which is a weighted sum of their in-vehicle time and wait time. $J_{in-vehicle,n}(v,\xi)$ is the in-vehicle travel time for passenger $n$ assigned to vehicle $v$ given route $\xi$, which is their drop-off time subtracted by their pickup time. $J_{wait,n} (v,\xi)$ is the wait time for passenger $n$ assigned to vehicle $v$ given route $\xi$, which is determined by subtracting the submission time from the pickup time. Wait time is assumed to be perceived more negatively for passengers than the in-vehicle time as reflected by parameter $\alpha$ for a multiplier over in-vehicle time cost. 

The episodes of experiences are extracted from the simulator outputs, which contain the vehicles' location and the number of requests assigned to each vehicle at each time step. In most simulations, a time step of either 30 seconds or 1 minute is used. The simulation updates after each time step and all information can be recorded with this level of detail. This time step might be too short for the definition of states explained in Section \ref{subsec:ProblemFormulation}. Longer periods, e.g. 5-minute time intervals, can be used for state definition by aggregating the simulation data. Similarly, the location of vehicles can be aggregated to zones for the state definition. Since only state and rewards are used for updating the state values, we can keep the needed information for policy evaluation and discard the rest. Therefore, an entire episode for a vehicle is represented by a list of (s, r) pairs. The data from various vehicles collectively contributes to learning a unified set of value functions. In other words, there are no individualized value functions for each vehicle. Instead, values are associated with each state characterized by the spatiotemporal status of the vehicle. The learned value functions are then used in the centralized dispatch unit in the online planning components.

\begin{figure}
    \centering
    \includegraphics[scale=0.5]{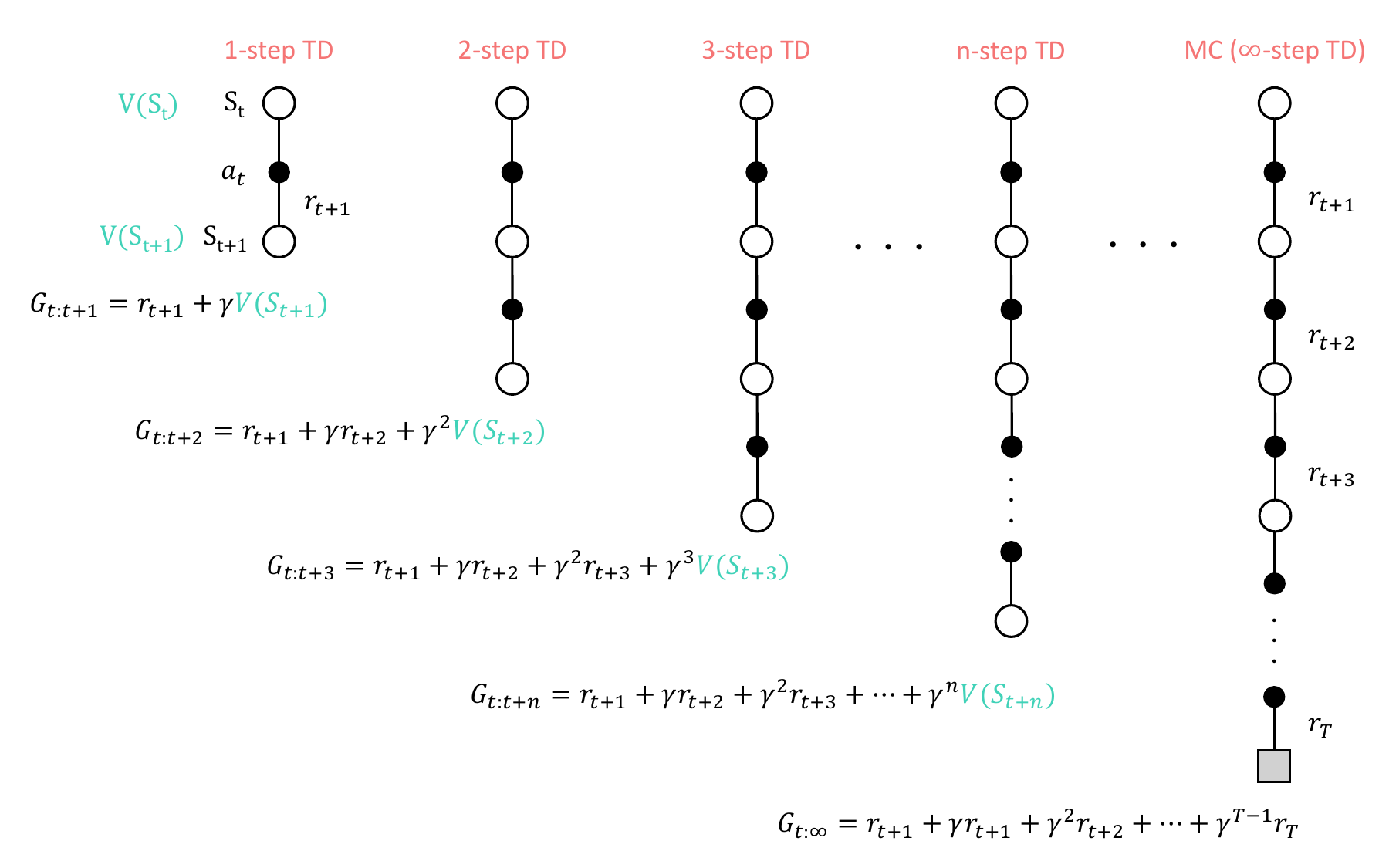}
    \caption{Comparison of updates in TD, n-step TD, and MC methods.}
    \label{fig:modelFree}
\end{figure}

Among the model-free methods for policy evaluation, Monte Carlo learning updates each state value based on the entire sequence of rewards observed from that state until the end of the episode. In contrast, the one-step TD method learns from incomplete episodes and updates the states based on only the next immediate reward, using the value of the subsequent state as an estimate for future rewards. An entire episode in a ride-pooling system concludes when the service operating hours end. However, it might be overly far-sighted and unnecessary for real-time dispatch decisions to factor in the end of operating hours at each time step. On the other hand, using a one-step TD target may introduce excessive bias in the learning process. A more balanced approach is the n-step TD method, which combines the TD and Monte Carlo methods by considering the observed rewards of $n$ steps ahead. Figure \ref{fig:modelFree} compares the updates in the three mentioned methods. The discounted return $G_{t:t+n}^v$ using n-step TD for vehicle $v$ at state $S_t$ is found in Eq. \ref{eq:n-stepReturn}.

\begin{equation}
    \label{eq:n-stepReturn}
    G_{t:t+n}^v = r_{t+1}^v + \gamma r_{t+2}^v + ... + \gamma^{n-1} r_{t+n}^v + \gamma^n V(S_{t+n})
\end{equation}

where $n$ is the number of future steps considered, $r_{t+i}^v$ is the immediate reward at time $t+i$ for vehicle $v$, $\gamma$ is the discount factor, and $V(S_{t+n})$ is the value of state $S_{t+n}$.

For example, looking one hour ahead and defining 5-minute time steps leads to $n=12$. For a vehicle in zone $z$ at time index $t$, $r_{t+1}^v$ would be the number of requests assigned to this vehicle in the first 5 minutes after t, $r_{t+2}^v$ is the number of assignments to the vehicle in the second 5 minutes after t, i.e. between 5 and 10 minutes after t, and so on. The vehicle could be moving around the network to pick up or drop off passengers after time index $t$  and end up at zone $z'$ at time index $t+12$. The value of this state $(z',t+12)$ is used to find the return at state $(z,t)$.

The initial value of all states is assumed to be zero. Each state value $S_t$ is updated according to Eq. \ref{eq:counterUpdate} and Eq. \ref{eq:valueUpdate} when a vehicle $v$ is in state $S_t$ and its return $G_t^v$ is determined. $N_{S_t}$ is a counter variable that keeps track of the number of instances used to update the value function $V{(S_t)}$. The offline learning algorithm is shown in Algorithm \ref{alg:offlineLearning}.

\begin{equation}
    \label{eq:counterUpdate}
    N(S_t) = N(S_t) + 1
\end{equation}

\begin{equation}
    \label{eq:valueUpdate}
    V(S_t) = V(S_t) + \frac{1}{N(S_t)} (G_t^v - V(S_t))
\end{equation}

\begin{algorithm}
\caption{Offline learning}
\label{alg:offlineLearning}
\begin{algorithmic}
\State Generate historical data using a simulator with a large fleet size
\State Initialize all state values $V(S)$ and state counters $N(S)$ to zero
\State Aggregate generated data to $|T|$ time intervals
\State Aggregate vehicle locations to $|Z|$ zones 
\For{each time index $t$ in $T$}
    \For{each vehicle $v$ in the system}
        \State \text{Find n-step return for vehicle $v$ at $S_t$:} $G_{S_t}^v = r_{t+1}^v + \gamma r_{t+2}^v + ... + \gamma^{n-1} r_{t+n}^v + \gamma^n V(S_{t+n})$
        \State \text{Increment counter:} $N(S_t) = N(S_t) + 1$
        \State \text{Update state value:} $V(S_t) = V(S_t) + \frac{1}{N(S_t)} (G_{S_t}^v - V(S_t))$
    \EndFor
\EndFor
\State Return a lookup table of states and their values $V(S)$
\end{algorithmic}
\end{algorithm}

\subsubsection{Online planning for matching} \label{subsec: OnlinePlanningMatching}
The online planning step for matching uses the learned value functions in the offline learning step to make real-time dispatch decisions. For each submitted request, the simulator finds a set of vehicles that meet the feasibility criteria explained in Section \ref{subsec:ProblemDescription} to serve the request. The central dispatch unit uses the online planning algorithm to find the best vehicle among the feasible ones. The system's objective is to optimize marginal expected gain, which is formulated as the objective function in Eq. \ref{eq:planningObjective} for each dispatch decision. 

\begin{equation}
    \label{eq:planningObjective}
    arg \max_v \left ( R_v + \gamma^{\Delta t_{S'_v}}V(S'_v) - \gamma^{\Delta t_{S_v}}V(S_v) \right )
\end{equation}

The value in parentheses represents the marginal expected gain of serving the request by vehicle $v$. The central dispatch unit chooses the vehicle that has the highest marginal expected gain. The objective function consists of two components: an immediate gain ($R_v$), which is explained later in this section, and a future gain, represented by the discounted difference in vehicle's state values before and after assigning the new request represented by $V(S_v)$ and $V(S'_v)$, respectively. 

\begin{figure}
    \centering
    \includegraphics[scale=0.8]{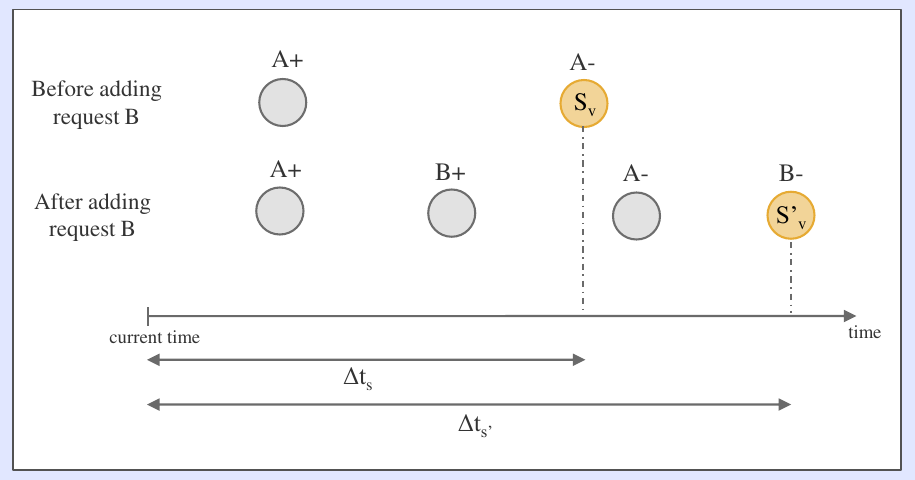}
    \caption{Definition of vehicle state in a ride-pooling system.}
    \label{fig:vehicleState}
\end{figure}

At any given time in the system, the vehicle can have multiple stops on its scheduled route. We define the vehicle's state by its final scheduled stop, i.e. the last drop-off location and time. For example, as shown in Figure \ref{fig:vehicleState}, the dispatch decision for request B is taking place at the current time in the system. Before assigning request B to vehicle $v$, the vehicle has two scheduled stops for picking up and dropping off passenger $A$ on its route. The vehicle's state $S_v$ in this case is defined by its final stop which is the dropoff location and time of request A. This time is occurring in the future, which means that the vehicle will be in this state in $\Delta t_s$ time periods from the current time. Therefore, the value of $S_v$ is discounted by $\gamma^{\Delta t_{S_v}}$ in Eq. \ref{eq:planningObjective}. After adding request B's pickup and dropoff locations to the scheduled route of vehicle $v$, its state $(S'_v)$ is similarly determined by its final stop, which is the dropoff location and time of request B. This state is happening $\Delta t_s'$ intervals from the current time; therefore, its value is discounted by $\gamma^{\Delta t_{S'_v}}$ in Eq. \ref{eq:planningObjective}. 

The future gain component tries to assign requests to vehicles that will position them in better states in the future compared to their current state. Thus, as discussed in \citet{xu2018large}, in similar situations, vehicles in lower-value states are more likely to be chosen, as their chances of being assigned to future requests are lower than those in higher-value states. Additionally, discounting the future state values gives more advantages to earlier drop-offs (and pickups indirectly) since there is more certainty about the near future. 

The future gain component takes the future state of vehicles into account. However, it does not consider the immediate impact of assigning a request to a vehicle, which includes the increases in the travel time of passengers already assigned to the vehicle (either onboard the vehicle or waiting for pickup), the travel time of the passenger who is about to be assigned to the vehicle, and the increase in the vehicle time traveled. These aspects are captured using the immediate gain component $(R_v)$ in the objective function in Eq. \ref{eq:planningObjective}, which is represented in Eq. \ref{eq:immediateGain}.

\begin{equation}
    \label{eq:immediateGain}
    R_v = \lambda \left ( c(v,\xi) - c(v,\xi') \right )
\end{equation}

where $c(v,\xi)$ is the current cost for (vehicle $v$, route $\xi$) as explained in Eq. \ref{eq:myopicPolicy} and $c(v,\xi')$ is the cost after assigning the new request to vehicle $v$. Since the cost values (in the immediate gain component) and the state values (in the future gain component) are in different orders of magnitude in Eq. \ref{eq:planningObjective}, parameter $\lambda$ is used to scale the cost values to match the magnitude of the state values, and is approximated using Eq. \ref{eq:lambda}.

\begin{equation}
    \label{eq:lambda}
    \lambda = \frac{\mathbb{E}\left({\gamma^{\Delta t_{s'_v}}V(s'_v) - \gamma^{\Delta t_{s_v}}V(s_v)} \right)}{\mathbb{E} \left( {c(v,\xi)-c(v,\xi')} \right)}
\end{equation}

%However, one potential drawback of discounted state values in the future gain component is that if a vehicle has no scheduled route and is currently in a high-value state, it may be more likely to stay idle and miss good potential matches.

\subsubsection{Online planning for rebalancing} \label{subsec: OnlinePlanningRebalancing}

Rebalancing operations are performed at intervals of $\tau$, which may differ from the system’s update interval. At each rebalancing time interval $\tau$, deficit zones (with supply lower than demand) and surplus zones (with supply exceeding demand) are identified, and idle vehicles are repositioned from surplus to deficit zones. The main challenges lie in representing demand in each zone and determining the optimal matches between surplus and deficit zones. In the offline learning step, demand is already represented using learned value functions, and the algorithm developed in the online planning step for matching can be applied to find the optimal matches between surplus and deficit zones as explained below.

The learned value functions in the offline learning step are used to represent the relative demand in each zone by dividing the zone's state value at a given time by the sum of all zones' state values at that specific time step as shown in Eq. \ref{eq:demand}, where $D_{(t,z)}$ represents the relative demand in zone $z$ at time $t$, and $V_{(t,z)}$ is the value of being in state (time $t$, zone $z$). The relative supply availability in each zone $z$ at time $t$ is defined as the number of vehicles located in that zone at time $t$ divided by the total number of vehicles in the system as shown in Eq. \ref{eq:supply}. Vehicles present in a zone may be en route to serve already assigned passengers; therefore, their availability for new requests within the zone is conditional on satisfying the constraints described in Section \ref{subsec:ProblemDescription}. For simplicity and scalability, however, the definition in Eq. \ref{eq:supply} assumes that all such vehicles are available to serve new requests within that zone. 

The difference between the relative demand and supply found using Eq. \ref{eq:demand} and \ref{eq:supply} is used to identify surplus and deficit zones in Eq. \ref{eq:zoneImbalance}. If a zone's imbalance value $\Delta_{(t,z)}$ is negative, the zone is in deficit; otherwise, it is in surplus. The objective is to reposition idle vehicles from surplus to deficit zones. Since surplus zones may not have sufficient idle vehicles, priority is given to deficit zones with larger imbalance values by sorting them based on $\Delta_{(t,z)}$. The algorithm then iterates through the ordered deficit zones, assigning the most suitable idle vehicle from the surplus zones according to the matching policy described in the previous section. For each candidate rebalancing request between a surplus zone and a deficit zone, the pickup location is randomly chosen within the surplus zone, and the drop-off location is randomly chosen within the deficit zone. Detailed steps are shown in Algorithm \ref{alg:onlineLearningRebalancing}.

\begin{equation}
    \label{eq:demand}
    D_{(t,z)} = \frac{V_{(t,z)}}{\sum_{z \in Z} V_{(t,z)}}   
\end{equation}

\begin{equation}
    \label{eq:supply}
    A_{(t,z)} = \frac{number\:of\:vehicles\:in\:zone\:z\:at\:time\:t}{fleet\:size}   
\end{equation}

\begin{equation}
    \label{eq:zoneImbalance}
    \Delta_{(t,z)} = A_{(t,z)}-D_{(t,z)}   
\end{equation}

\begin{algorithm}
\caption{Online planning for rebalancing}
\label{alg:onlineLearningRebalancing}
\begin{algorithmic}
\State Specify rebalancing time interval $\tau$

\For{each rebalancing time interval $t$}
    \State Initialize $Z_{deficit}$ and $Z_{surplus}$ as empty sets
    \State \text{Find demand in each zone at time $t$}: $D_{(t,z)} = \dfrac{V_{(t,z)}}{\sum_{z \in Z}V_{(t,z)}}$
    \State \text{Find supply in each zone at time $t$}: $A_{(t,z)} = \dfrac{\text{number of vehicles in zone $z$ at time $t$}}{\text{fleet size}}$
    \State \text{Find each zone's imbalance at time $t$}: $\Delta_{(t,z)} = A_{(t,z)}-D_{(t,z)}$
    \If{$\Delta_{(t,z)}<0$} 
        \State Add $z$ to $Z_{deficit}$
    \Else
        \State Add $z$ to $Z_{surplus}$
    \EndIf
     \State Find idle vehicles located in surplus zones: $V_{idle}^{s}$
     \State Sort $Z_{deficit}$ in ascending order based on value of $\Delta_{(t,z)}$
    \For{each zone $z$ in $Z_{deficit}$}
        \State Find best vehicle in $V_{idle}^s$ to reposition to $z$ using the matching policy
        \State Remove the best found vehicle from $V_{idle}^s$
    \EndFor
\EndFor
\end{algorithmic}
\end{algorithm}
\section{Computational experiments} \label{sec:ComputationalExperiments}

We modify the simulator developed by \citep{namdarpour2024non} to implement our proposed method in a simulation environment and evaluate its performance using NYC taxi data \citep{NYCTaxiData} \footnote{The code is available at \url{https://github.com/BUILTNYU/nomad-rps}}. The simulation setup and the results for both matching and rebalancing are presented in separate sections below.

\subsection{Simulation setup}
\label{subsec:SimulationSetup}
\begin{figure}
\centering
    \begin{minipage}{.45\textwidth}
        \centering
        \includegraphics[scale=0.3]{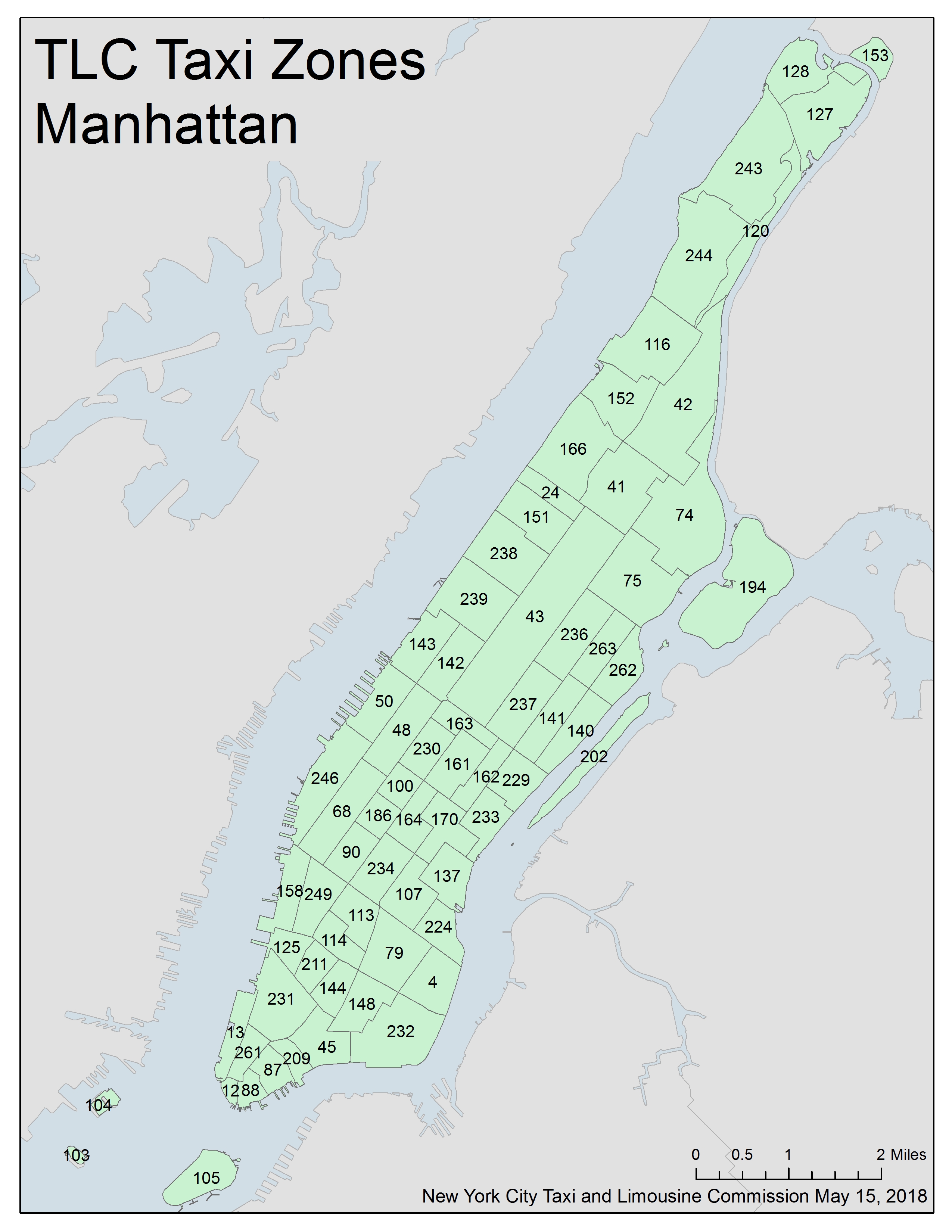}
        \caption{Taxi zones in Manhattan \citep{NYCTaxiData}.}
        \label{fig:ManhattanTaxiZones}
    \end{minipage}%
    \begin{minipage}{.45\textwidth}
        \centering
        \includegraphics[scale=0.3]{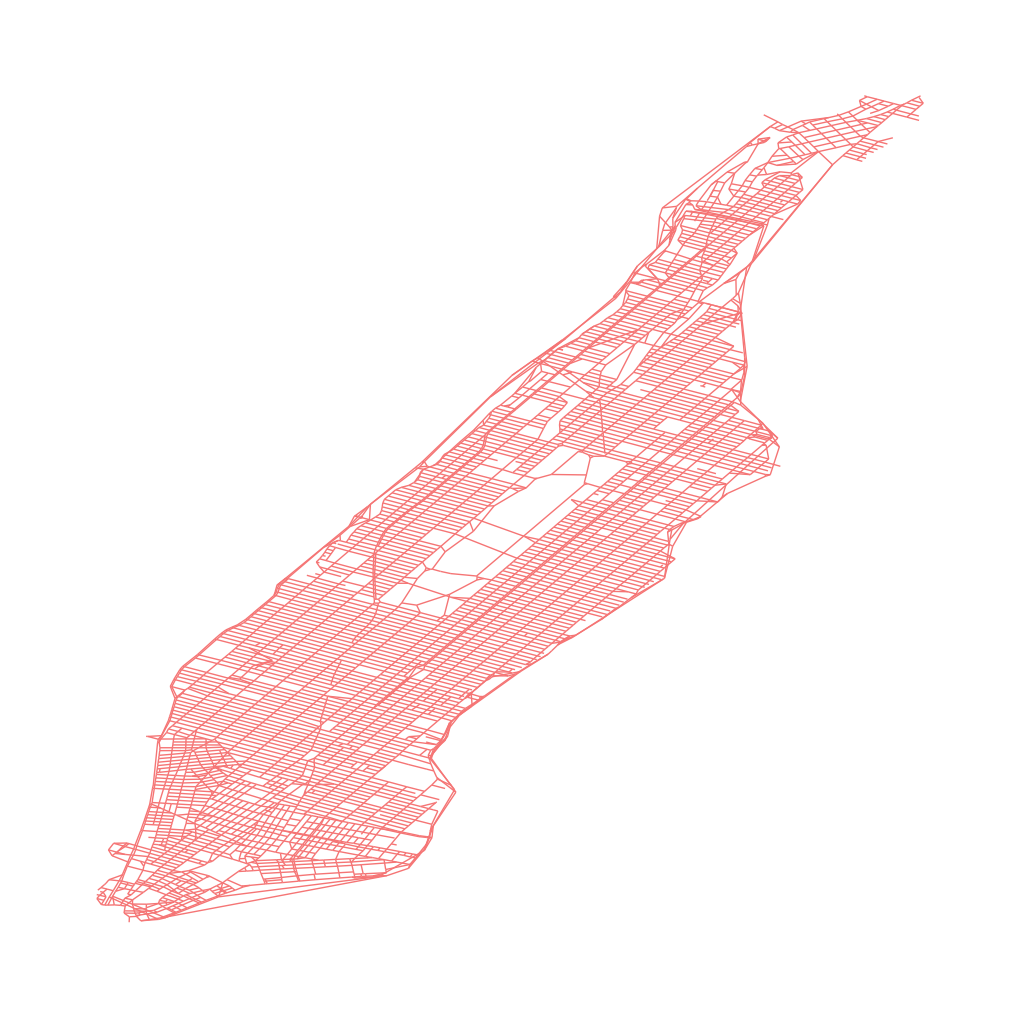}
        \caption{Manhattan network.}
        \label{fig:ManhattanNetwork}
    \end{minipage}
\end{figure}

The NYC taxi data for the month of February 2024 was used for the simulation. After excluding weekends and public holidays, we ended up with 20 days of data, using the first 14 days for offline learning and the remaining 6 days for testing the online planning algorithm. The exact dates of data are as follows:
\begin{itemize}
    \item 14-day learning set: 1st-2nd, 5th-9th, 12th-16th, 20th-21st February 2024
    \item 6-day testing set: 22nd-23rd, 26th-29th February 2024
\end{itemize}
The dataset includes pickup and drop-off times and locations, represented by taxi zone IDs. We assumed the pickup time represents the request submission time and that all requests correspond to single passengers. We only considered trips with both pickup and drop-off locations within Manhattan, where there are 69 taxi zones as shown in Figure \ref{fig:ManhattanTaxiZones}. On average, there are 88,039 requests per day. 

We used the same road network for Manhattan as used in \citet{alonso2017demand}. This network has 4,091 nodes and 9,452 links as shown in Figure \ref{fig:ManhattanNetwork}. The weight of each link is its daily mean travel time estimated by the method proposed by \citet{santi2014quantifying}. 

Equal weights were used in the simulator for the operator’s cost and passengers’ cost in the cost function shown in Eq. \ref{eq:myopicPolicy} ($\theta = 0.5$).  According to \citet{yap2023ride}, waiting time for ride-hailing services is perceived about 1.4 times more negatively than in-vehicle time. Therefore, a coefficient of 1.4 is chosen for $\alpha$ in Eq. \ref{eq:myopicPolicy}. The vehicle passenger capacity was assumed 6, excluding the driver. The maximum wait time for passengers was set to 10 minutes. The time step in the simulation is 30 seconds, meaning that the states and rewards are also recorded every 30 seconds. The rebalancing time interval $\tau$ is also set to 30 seconds. For the latest drop-off time and the dwell time, the same parameters are used as in \citet{namdarpour2024non}. The initial location of vehicles is selected randomly from the network nodes. 

For the state space $S$, 5-minute intervals were defined for a 24 hour period, i.e. $|T|=288$, and the taxi zones were used as the zone indices, i.e. $|Z|=69$. Therefore, the size of the state space is $288 \times 69 = 19,872$. A large fleet size of 7,000 vehicles was used in the simulator to serve the historical demand data for each day, ensuring that no requests were rejected and all demand was captured in the learning process. This fleet size was determined by testing various sizes and selecting the one that resulted in no rejections. The 12-step TD learning method was used for the offline learning step with a discount factor $\gamma$ of 0.9. The offline learning phase was conducted on 14 days of data, with each day's learning taking less than 3 minutes. The state values were updated based on new data, building upon the values from previous days. The final state value lookup table was then used for the online planning phases. Based on the obtained data, a value of 0.005 was used for $\lambda$ in Eq. \ref{eq:lambda}.

\subsection{Matching results}
\label{subsec:matchingResults}

To evaluate the performance of our proposed method for matching, the results were compared to two other matching algorithms across different fleet sizes using the 6-day testing data explained in Section \ref{subsec:SimulationSetup}. The tested algorithms are explained below. Vehicle repositioning is not considered in this section.

\begin{itemize}
    \item \textbf{Myopic}: The myopic policy explained in Eq. \ref{eq:myopicPolicy} used as the baseline.

    \item \textbf{NM-beta}: The CFA policy proposed by \citet{hyytia2012non} which has a tunable parameter for lookahead approximation represented by $\beta$.

    \item \textbf{NM-RL}: The proposed learning and planning framework in this study that uses RL to optimize the system over a long-term horizon in real time for matching decisions.
\end{itemize}

\begin{figure}
    \centering
    \includegraphics[scale=0.59]{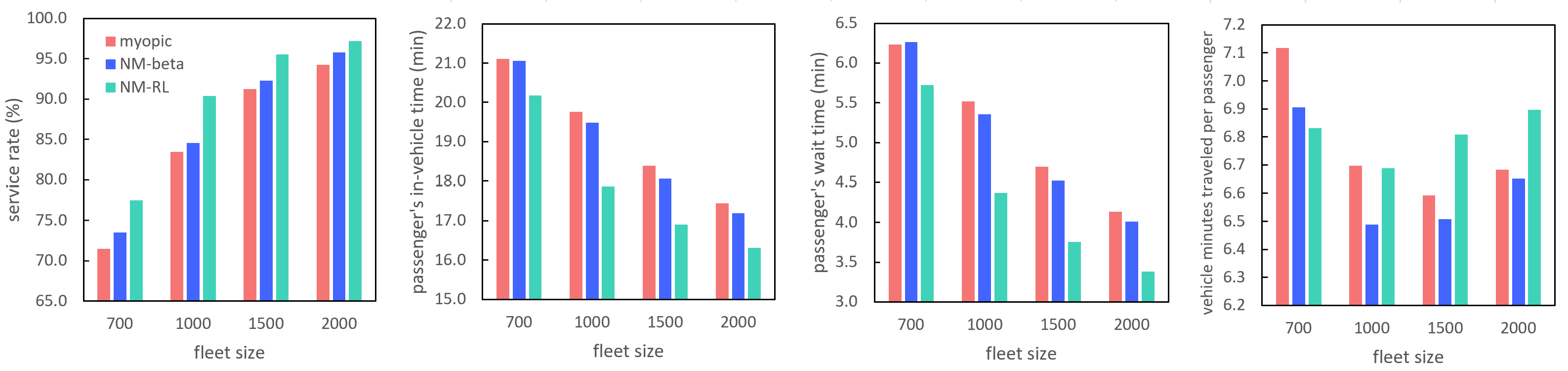}
    \caption{Comparison of test results for myopic, NM-beta, and NM-RL methods across different fleet sizes.}
    \label{fig:testResults}
\end{figure}

For the NM-beta algorithm, the value of $\beta$ was calibrated using a grid search method and the first 14 days of data (see Section \ref{subsec:SimulationSetup} for exact dates) with a fleet size of 2000. The calibrated value is 0.005. The results of testing the three algorithms on the remaining 6 days of data are shown in Figure \ref{fig:testResults} for different fleet sizes, including 700, 1000, 1500, and 2000 vehicles in the system. From the passengers' perspective, the two measures of in-vehicle time (drop-off time subtracted by the pickup time) and wait time (pickup time subtracted by the request submission time) are selected. From the operator's perspective, the two measures of service rate (the percentage of accepted requests), and vehicle minutes traveled per passenger (VMT) are selected. In general, NM-RL achieves the highest service rate across different fleet sizes and leads to the shortest in-vehicle and wait time for passengers among the studied algorithms at the cost of a slight increase in VMT. Each measure is explained in more detail below.

\textbf{Service rate}: The results in Figure \ref{fig:testResults} show that NM-beta achieves better results than the myopic approach, while the NM-RL algorithm achieves the highest service rate for all fleet sizes. It is evident that using RL is significantly more effective than the CFA policy employed in NM-beta for improving the service rate.  NM-RL results in the highest increase of +8.4\% compared to the myopic approach for the smallest fleet size, with the increase gradually reducing to +3.1\% as the fleet size grows to 2000. The results demonstrate that a myopic approach exacerbates service performance for smaller fleet sizes, indicating substantial potential for improvement with a non-myopic, RL-based approach. 

\begin{figure}
    \centering
    \includegraphics[scale=0.4]{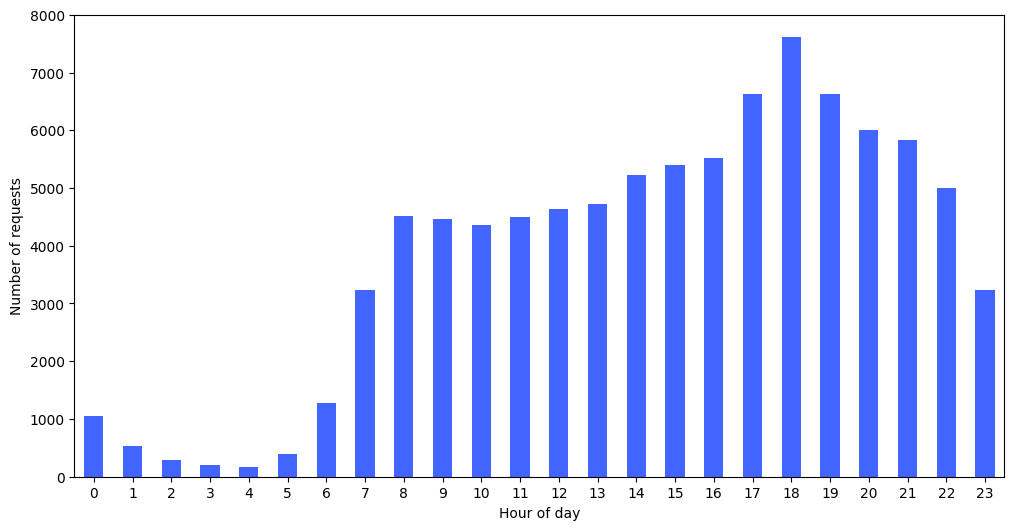}
    \caption{Distribution of submitted requests during different hours of day averaged over 6-day test dataset.}
    \label{fig:RequestTimeDist}
\end{figure}

\begin{figure}
    \centering
    \includegraphics[scale=0.9]{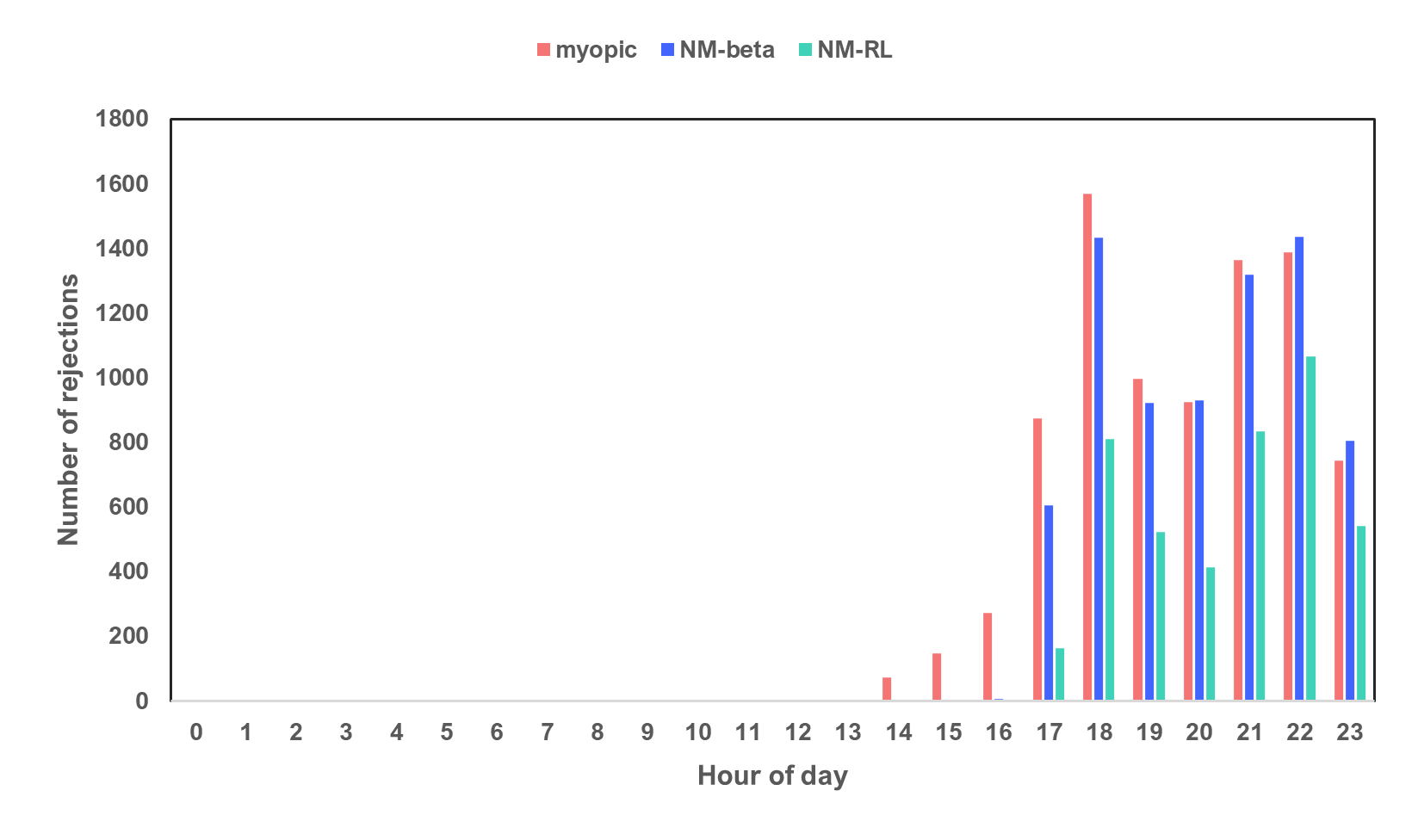}
    \caption{Distribution of rejections during different hours of day averaged over 6-day test dataset using different dispatch methods.}
    \label{fig:RejectionsTimeDist}
\end{figure}

\textbf{Rejections}: The distribution of submitted requests and number of rejections using different dispatch methods for different hours of day using a fleet size of 1500 are shown in Figure \ref{fig:RequestTimeDist} and \ref{fig:RejectionsTimeDist}, respectively. The values are found by averaging over 6 days of testing data. As shown in both figures, the peak of the demand occurs at 6 PM, which is also when rejections peak using a myopic method. While the NM-beta method reduces the rejections during most hours, it can be seen that NM-RL has a significantly better performance during all hours and reducing the rejections by around 50 percent at the peak hour. The improvement percentage for NM-RL compared to the myopic approach decreases in the final hours of the day. This can be attributed to the fact that no requests are considered after 24 hours, making the 1-hour lookahead in the non-myopic approach less effective during these later hours.

\textbf{Passengers' in-vehicle and wait times}: As seen in Figure \ref{fig:testResults}, both in-vehicle time and wait time for passengers are significantly improved using the NM-RL method compared to the myopic approach, ranging from 9.6\% and 20.8\% decrease in in-vehicle time and wait time, respectively, for fleet size of 1000 to 4.4\% and 8.2\% decrease for fleet size of 700. This suggests that defining the rewards as the number of assigned requests to vehicles could indirectly improve passengers' travel time by navigating vehicles to locations with higher state values. This impact is observed more significantly in the reduced passenger wait times, demonstrating that vehicles are effectively positioned close to where demand arises. 

\textbf{Vehicle minutes traveled per passenger (VMT)}: For this measure, it can be seen that NM-beta has the best performance by reducing VMT across all fleet sizes. NM-RL reduces the VMT by 4.0\% and 0.1\% for smaller fleet sizes of 700 and 1000, respectively, compared to the myopic approach. However, it increases the VMT by 3.3\% and 3.2\% for 1500 and 2000 fleet sizes. This observation can be explained by the system assigning vehicles to requests with higher destination values, potentially resulting in overall longer VMTs. However, these increases can be justified by the substantial improvements observed in the other three measures. 

\textbf{Fleet size reduction}: Additionally, it can be seen that by reducing the fleet size from 2000 to 1500, NM-RL still achieves a better service rate of 95.5\% compared to the myopic approach with a fleet size of 2000 (94.3\%), while passengers also experience shorter in-vehicle and wait times. This suggests the possibility of reducing the fleet size by over 25\% while maintaining the same level of performance, translating to significant cost savings for the operator. 

\begin{figure}
\centering
\begin{subfigure}{.3\textwidth}
  \centering
  \includegraphics[scale=0.3]{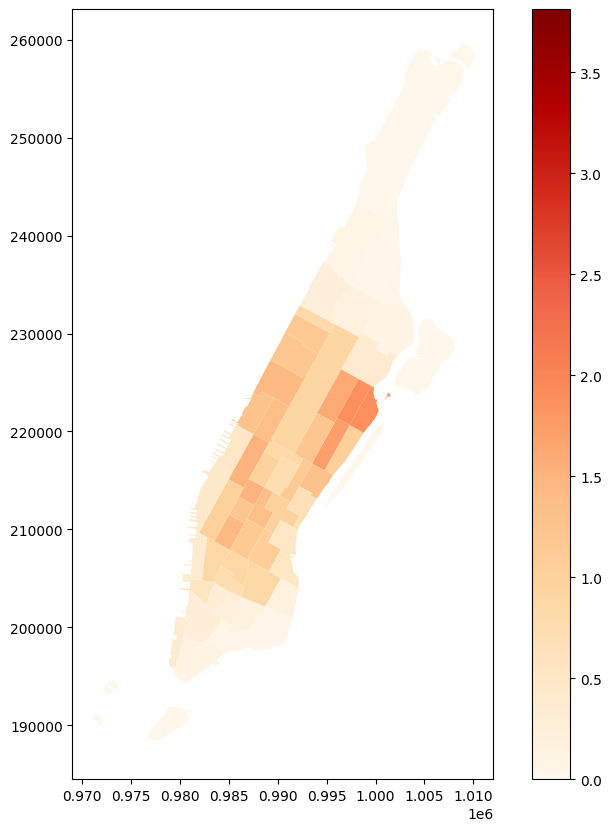}
  \caption{8 AM}
\end{subfigure}%
\begin{subfigure}{.3\textwidth}
  \centering
  \includegraphics[scale=0.3]{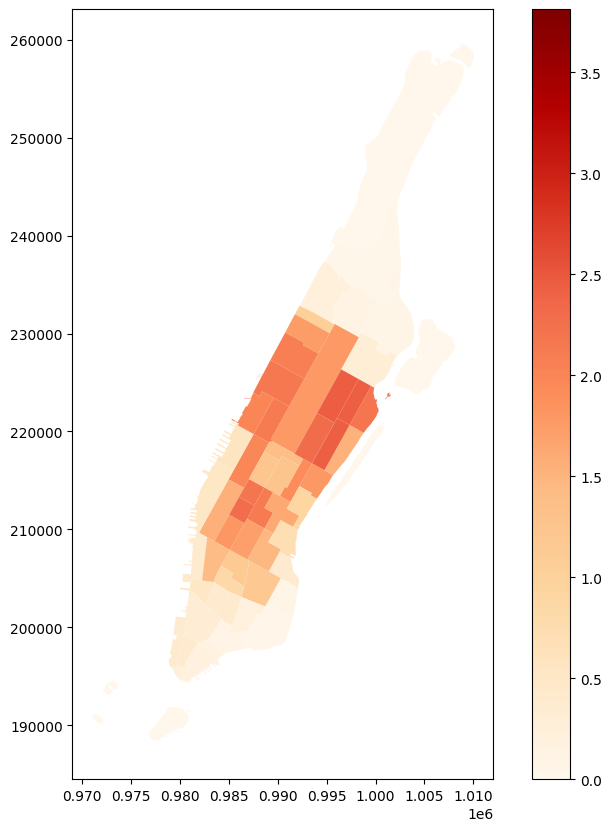}
  \caption{1 PM}
\end{subfigure}
\begin{subfigure}{.3\textwidth}
  \centering
  \includegraphics[scale=0.3]{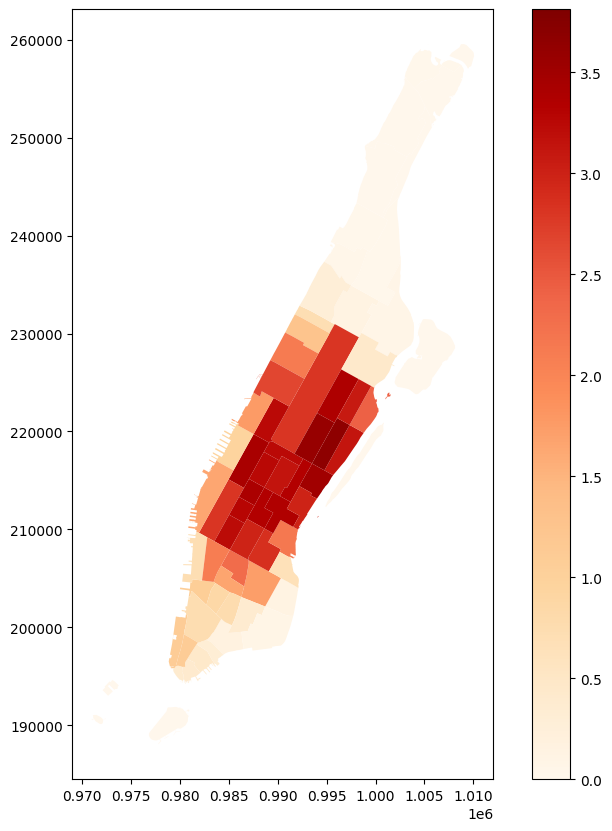}
  \caption{6 PM}
\end{subfigure}
\caption{Visualization of learned state values in Manhattan at (a) 8 AM (b) 1 PM and (c) 6 PM}
\label{fig:StateValuesVisualization}
\end{figure}

\textbf{Visualization of state values}: The final state values after training on 14 days of data are shown as zonal heatmaps in Figure \ref{fig:StateValuesVisualization} for different hours of day, including 8 AM, 1 PM, and 6 PM, representing morning, mid-day, and evening hours. As shown in Figure \ref{fig:RequestTimeDist}, submissions are at their peak at 6 PM. This point can be seen in Figure \ref{fig:StateValuesVisualization} as the state values take higher values at 6 PM compared to 8 AM and 1 PM. As the evening peak marks the end of working hours, it's not surprising to see many valuable zones in Midtown, which is a commercial district. Zones around central park are also valuable areas which could be the origin of personal and recreational trips. In the morning, high value zones are mostly concentrated in residential areas, and similar patterns can be seen during mid-day while the state values are generally higher as the demand follows an increasing trend after 1 PM compared to 8 AM (see Figure \ref{fig:RequestTimeDist}).

\begin{figure}
    \centering
    \includegraphics[scale=0.66]{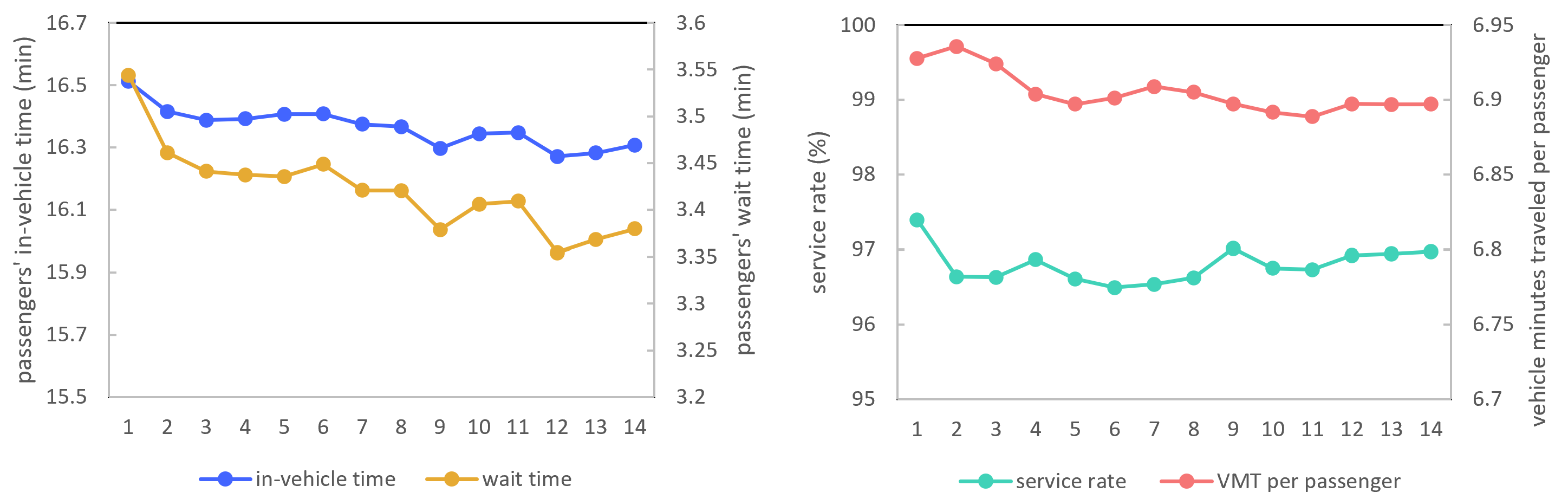}
    \caption{The impact of training set size on performance measures.}
    \label{fig:trainingSize}
\end{figure}

\textbf{Effect of training set size}: As mentioned earlier, results were obtained by learning the state values on a 14-day dataset. In this section, the effect of training set size is investigated by incrementally increasing the dataset size from 1 day to 14 days, and similarly, evaluating the impact on the 6-day dataset in the online planning phase. Results are shown in Figure \ref{fig:trainingSize}. As can be seen, the service rate achieves high performance even with 1 day of learning. However, other performance measures improve by increasing the size of the training set. This suggests that while the system can achieve a high service rate with a smaller training set, the system learns to dispatch vehicles more efficiently by training on a larger dataset. This results in vehicles being closer to pickup locations and reducing passengers' wait time, in-vehicle time, and VMT per passenger. Since the same state values are used for different working days, some levels of oscillation can be seen in the graphs, as the demand patterns are not exactly the same across different days. However, all measures reach a steady state after approximately 12 days of learning.

\begin{figure}
\centering
\begin{subfigure}{.4\textwidth}
  \centering
  \includegraphics[scale=0.3]{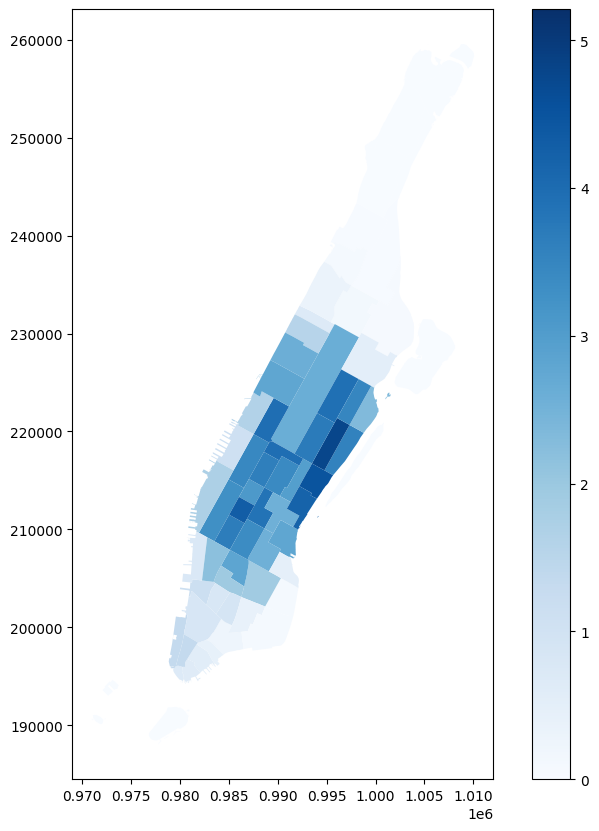}
  \caption{1-day training set}
\end{subfigure}%
\begin{subfigure}{.4\textwidth}
  \centering
  \includegraphics[scale=0.3]{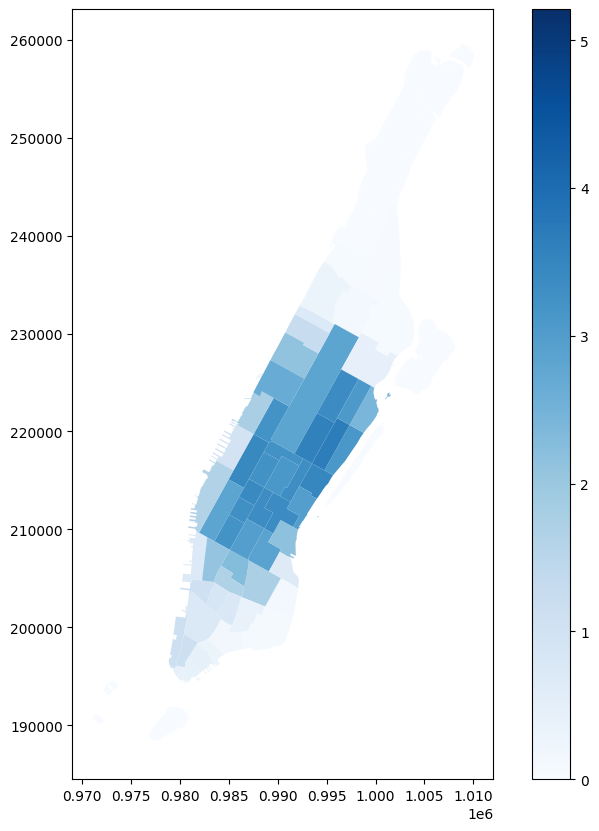}
  \caption{14-day training set}
\end{subfigure}
\caption{Visualization of learned state values in Manhattan at 6 PM after training on (a) 1 day of data and (b) 14 days of data}
\label{fig:StateValuesDataSize}
\end{figure}

The impact of training set size on learned state values is visualized in Figure \ref{fig:StateValuesDataSize} for 6 PM using 1 day and 14 days of data. By learning the value functions using a larger dataset, the values become smoother as it captures patterns across multiple days and reduces the spikes caused by learning from a single day. The largest state value in the first case is 5.21 while in the second case it is reduced to 3.81. Please note that Figure \ref{fig:StateValuesDataSize} (b) shows the same values as Figure \ref{fig:StateValuesVisualization} (c), but uses a different color scale to highlight the maximum value for the 1-day training set in Figure \ref{fig:StateValuesDataSize} (a).

\subsection{Rebalancing results}
\label{subsec:rebalancingResults}

\begin{figure}
    \centering
    \includegraphics[scale=1.3]{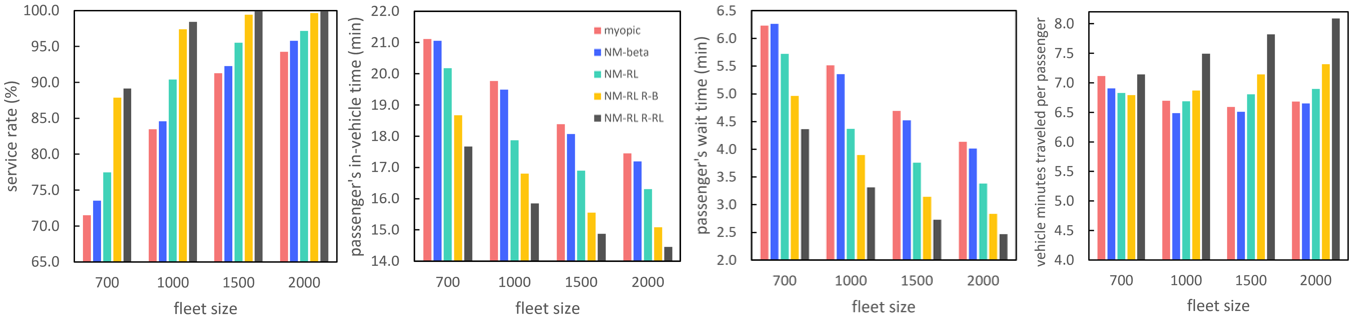}
    \caption{Comparison of test results for myopic, NM-beta, NM-RL, NM-RL R-B, NM-RL R-RL methods across different fleet sizes.}
    \label{fig:R-testResults}
\end{figure}

To evaluate the proposed framework for rebalancing operations, we compare our results with the simple yet effective rebalancing method proposed by \citet{alonso2017demand}. In their approach, whenever a request is rejected, a rebalancing request is generated to dispatch an idle vehicle to the pickup location of the rejected request. If no idle vehicle is available at that moment, the rebalancing request is ignored. We apply our proposed matching algorithm to select the most suitable idle vehicle for dispatch to implement their method. Both rebalancing approaches -- the method proposed by \citet{alonso2017demand}, denoted as \textit{R-B}, and our proposed method, denoted as \textit{R-RL} -- are incorporated into the proposed non-myopic matching framework. Their performance is evaluated using the same metrics in the prevous section and presented in Figure \ref{fig:R-testResults} alongside the three baseline matching algorithms without rebalancing.

\begin{itemize}
    \item \textbf{NM-RL R-B}: The rebalancing operation proposed by \citet{alonso2017demand} is added to the NM-RL matching.
    \item \textbf{NM-RL R-RL}: The RL-based rebalancing operation proposed in this study added to the NM-RL matching.
\end{itemize}

For brevity, the prefix \textit{NM-RL} is omitted from the algorithm names in the following discussion. Results in Figure \ref{fig:R-testResults} show that regardless of the algorithm used for rebalancing, repositioning vehicles can significantly improve passenger wait time and in-vehicle time, and service rate. Similar to matching results, the largest improvements are observed for smaller fleet sizes, with the benefits diminishing as fleet size increases. Compared to the NM-RL algorithm, where the proposed framework is used for matching but no repositioning is incorporated, R-RL and R-B increase service rate by up to 15.1\% and 13.4\%, respectively. R-RL reduces passenger in-vehicle time by up to 12.5\%, while R-B achieves a reduction of up to 8.0\%. The most notable improvement using our proposed algorithm is seen in passenger wait time: R-RL reduces it by up to 27.3\% while R-B by up to 16.3\%. Theses results demonstrate that the proposed algorithm effectively navigates vehicles toward areas of demand. While R-B reacts only after a rejection occurs, our method acts proactively before rejections take place. 

The cost of improvements achieved by rebalancing is reflected in vehicle minutes traveled (VMT) per passenger: R-RL increases VMT by up to 17.3\%, while R-B increases it by 6.1\%, thereby outperforming our method in this measure. However, in our algorithm this trade-off can be controlled by adjusting the rebalancing time interval $\tau$. Since matching decisions are made every 30 seconds, and R-B therefore can operate at the same frequency, we set the rebalancing interval of our proposed method to 30 seconds for comparability. In practice, however, rebalancing operations are typically less frequent, and the elevated VMT observed in our method reflects the impact of this high rebalancing frequency.
\section{Conclusion}
\label{sec:Conclusion}

This study proposes a learning and planning approach to make real-time non-myopic dispatch decisions, including both matching and rebalancing, by a centralized dispatch unit in a large-scale ride-pooling system. The framework includes an offline simulation-informed policy evaluation component and two online planning components. An n-step TD learning method is used for offline policy evaluation to learn spatiotemporal value functions from historical demand data fed into a simulator with a specified fleet size hyperparameter. The online learning components use the learned value functions and immediate rewards to make real-time decisions. The proposed methodology is implemented in a simulation platform and tested using real-world NYC taxi demand data \citep{NYCTaxiData}. Matching results are compared with a myopic and a CFA policy. The findings indicate that the proposed non-myopic approach can effectively capture the long-term consequences of matching decisions, improving service from both operators' and users' perspectives. The service rate increases by up to 8.4$\%$ compared to a myopic policy while passenger wait time and in-vehicle time are significantly decreased. This demonstrates that the system navigates vehicles to locations with higher state values, effectively capturing the supply and demand patterns. Additionally, the proposed methodology can reduce fleet size by more than 25$\%$ compared to a myopic policy, while maintaining the same level of performance, thereby offering significant cost savings for the operators. Incorporating rebalancing operations into our proposed framework reduces passenger wait time and in-vehicle time by up to 27.3\% and 12.5\%, respectively, while increasing the service rate by up to 15.1\%, compared to using our methodology for matching decisions alone. The proposed methodology does not account for electric charging operations, which can be explored in future work. Further research could also examine policy iteration methods to enhance performance and explore deeper integration of matching and rebalancing operations to improve vehicle minutes traveled per passenger.

\section*{Acknowledgements}
This work was supported by the C2SMARTER Center.  Earlier versions of this research were presented at the Transportation Research Board Annual Meeting 2025 and TRISTAN XII 2025.

\bibliographystyle{apalike}
\clearpage
\bibliography{library}

\begin{thebibliography}{}

\bibitem[Al-Abbasi et~al., 2019]{al2019deeppool}
Al-Abbasi, A.~O., Ghosh, A., and Aggarwal, V. (2019).
\newblock Deeppool: Distributed model-free algorithm for ride-sharing using deep reinforcement learning.
\newblock {\em IEEE Transactions on Intelligent Transportation Systems}, 20(12):4714--4727.

\bibitem[Alonso-Mora et~al., 2017]{alonso2017demand}
Alonso-Mora, J., Samaranayake, S., Wallar, A., Frazzoli, E., and Rus, D. (2017).
\newblock On-demand high-capacity ride-sharing via dynamic trip-vehicle assignment.
\newblock {\em Proceedings of the National Academy of Sciences}, 114(3):462--467.

\bibitem[Grab, 2024]{Grab}
Grab (2024).
\newblock Grabshare.
\newblock \url{https://www.grab.com/sg/transport/} [Accessed on June, 2024].

\bibitem[Gu{\'e}riau and Dusparic, 2018]{gueriau2018samod}
Gu{\'e}riau, M. and Dusparic, I. (2018).
\newblock Samod: Shared autonomous mobility-on-demand using decentralized reinforcement learning.
\newblock In {\em 2018 21st International Conference on Intelligent Transportation Systems (ITSC)}, pages 1558--1563. IEEE.

\bibitem[Haliem et~al., 2021]{haliem2021adapool}
Haliem, M., Aggarwal, V., and Bhargava, B. (2021).
\newblock Adapool: A diurnal-adaptive fleet management framework using model-free deep reinforcement learning and change point detection.
\newblock {\em IEEE Transactions on Intelligent Transportation Systems}, 23(3):2471--2481.

\bibitem[Han et~al., 2022]{han2022physics}
Han, Y., Wang, M., Li, L., Roncoli, C., Gao, J., and Liu, P. (2022).
\newblock A physics-informed reinforcement learning-based strategy for local and coordinated ramp metering.
\newblock {\em Transportation Research Part C: Emerging Technologies}, 137:103584.

\bibitem[Ho et~al., 2018]{ho2018survey}
Ho, S.~C., Szeto, W.~Y., Kuo, Y.-H., Leung, J.~M., Petering, M., and Tou, T.~W. (2018).
\newblock A survey of dial-a-ride problems: Literature review and recent developments.
\newblock {\em Transportation Research Part B: Methodological}, 111:395--421.

\bibitem[Hyyti{\"a} et~al., 2012]{hyytia2012non}
Hyyti{\"a}, E., Penttinen, A., and Sulonen, R. (2012).
\newblock Non-myopic vehicle and route selection in dynamic darp with travel time and workload objectives.
\newblock {\em Computers \& Operations Research}, 39(12):3021--3030.

\bibitem[Jindal et~al., 2018]{jindal2018optimizing}
Jindal, I., Qin, Z.~T., Chen, X., Nokleby, M., and Ye, J. (2018).
\newblock Optimizing taxi carpool policies via reinforcement learning and spatio-temporal mining.
\newblock In {\em 2018 IEEE International Conference on Big Data (Big Data)}, pages 1417--1426. IEEE.

\bibitem[Jung et~al., 2016]{jung2016dynamic}
Jung, J., Jayakrishnan, R., and Park, J.~Y. (2016).
\newblock Dynamic shared-taxi dispatch algorithm with hybrid-simulated annealing.
\newblock {\em Computer-Aided Civil and Infrastructure Engineering}, 31(4):275--291.

\bibitem[Li et~al., 2022]{li2022value}
Li, C., Parker, D., and Hao, Q. (2022).
\newblock A value-based dynamic learning approach for vehicle dispatch in ride-sharing.
\newblock In {\em 2022 IEEE/RSJ International Conference on Intelligent Robots and Systems (IROS)}, pages 11388--11395. IEEE.

\bibitem[Liu et~al., 2024]{liu2024joint}
Liu, Z., Ouyang, G., Zhang, B., Du, B., Chen, C., and Wu, K. (2024).
\newblock Joint order dispatching and vehicle repositioning for dynamic ridesharing.
\newblock {\em IEEE Transactions on Mobile Computing}.

\bibitem[Ma et~al., 2019]{ma2019dynamic}
Ma, T.-Y., Rasulkhani, S., Chow, J.~Y., and Klein, S. (2019).
\newblock A dynamic ridesharing dispatch and idle vehicle repositioning strategy with integrated transit transfers.
\newblock {\em Transportation Research Part E: Logistics and Transportation Review}, 128:417--442.

\bibitem[MOIA, 2024]{MOIA}
MOIA (2024).
\newblock Moia.
\newblock \url{https://www.moia.io/en} [Accessed on June, 2024].

\bibitem[Namdarpour et~al., 2024]{namdarpour2024non}
Namdarpour, F., Liu, B., Kuehnel, N., Zwick, F., and Chow, J.~Y. (2024).
\newblock On non-myopic internal transfers in large-scale ride-pooling systems.
\newblock {\em Transportation Research Part C: Emerging Technologies}, 162:104597.

\bibitem[Powell, 2019]{powell2019unified}
Powell, W.~B. (2019).
\newblock A unified framework for stochastic optimization.
\newblock {\em European Journal of Operational Research}, 275(3):795--821.

\bibitem[Qin et~al., 2022]{qin2022reinforcement}
Qin, Z.~T., Zhu, H., and Ye, J. (2022).
\newblock Reinforcement learning for ridesharing: An extended survey.
\newblock {\em Transportation Research Part C: Emerging Technologies}, 144:103852.

\bibitem[Santi et~al., 2014]{santi2014quantifying}
Santi, P., Resta, G., Szell, M., Sobolevsky, S., Strogatz, S.~H., and Ratti, C. (2014).
\newblock Quantifying the benefits of vehicle pooling with shareability networks.
\newblock {\em Proceedings of the National Academy of Sciences}, 111(37):13290--13294.

\bibitem[Sayarshad and Chow, 2015]{sayarshad2015scalable}
Sayarshad, H.~R. and Chow, J.~Y. (2015).
\newblock A scalable non-myopic dynamic dial-a-ride and pricing problem.
\newblock {\em Transportation Research Part B: Methodological}, 81:539--554.

\bibitem[Shah et~al., 2020]{shah2020neural}
Shah, S., Lowalekar, M., and Varakantham, P. (2020).
\newblock Neural approximate dynamic programming for on-demand ride-pooling.
\newblock In {\em Proceedings of the AAAI Conference on Artificial Intelligence}, volume~34, pages 507--515.

\bibitem[Shi et~al., 2021]{shi2021physics}
Shi, R., Mo, Z., Huang, K., Di, X., and Du, Q. (2021).
\newblock A physics-informed deep learning paradigm for traffic state and fundamental diagram estimation.
\newblock {\em IEEE Transactions on Intelligent Transportation Systems}, 23(8):11688--11698.

\bibitem[Singh et~al., 2021]{singh2021distributed}
Singh, A., Al-Abbasi, A.~O., and Aggarwal, V. (2021).
\newblock A distributed model-free algorithm for multi-hop ride-sharing using deep reinforcement learning.
\newblock {\em IEEE Transactions on Intelligent Transportation Systems}, 23(7):8595--8605.

\bibitem[Sutton and Barto, 2018]{sutton2018reinforcement}
Sutton, R.~S. and Barto, A.~G. (2018).
\newblock {\em Reinforcement learning: An introduction}.
\newblock MIT press.

\bibitem[Tang et~al., 2021]{tang2021value}
Tang, X., Zhang, F., Qin, Z., Wang, Y., Shi, D., Song, B., Tong, Y., Zhu, H., and Ye, J. (2021).
\newblock Value function is all you need: A unified learning framework for ride hailing platforms.
\newblock In {\em Proceedings of the 27th ACM SIGKDD Conference on Knowledge Discovery \& Data Mining}, pages 3605--3615.

\bibitem[Taxi and Commission, 2024]{NYCTaxiData}
Taxi, N. and Commission, L. (2024).
\newblock Tlc trip record data.
\newblock \url{https://www.grab.com/sg/transport/} [Accessed on June, 2024].

\bibitem[Uber, 2024]{Uber}
Uber (2024).
\newblock Uberx share.
\newblock \url{https://www.uber.com/us/en/ride/uberx-share} [Accessed on June, 2024].

\bibitem[Via, 2024]{Via}
Via (2024).
\newblock Via microtransit.
\newblock \url{https://ridewithvia.com/solutions/microtransit} [Accessed on June, 2024].

\bibitem[Wang et~al., 2023]{wang2023optimization}
Wang, D., Wang, Q., Yin, Y., and Cheng, T. (2023).
\newblock Optimization of ride-sharing with passenger transfer via deep reinforcement learning.
\newblock {\em Transportation Research Part E: Logistics and Transportation Review}, 172:103080.

\bibitem[Xu et~al., 2018]{xu2018large}
Xu, Z., Li, Z., Guan, Q., Zhang, D., Li, Q., Nan, J., Liu, C., Bian, W., and Ye, J. (2018).
\newblock Large-scale order dispatch in on-demand ride-hailing platforms: A learning and planning approach.
\newblock In {\em Proceedings of the 24th ACM SIGKDD international conference on knowledge discovery \& data mining}, pages 905--913.

\bibitem[Yap and Cats, 2023]{yap2023ride}
Yap, M. and Cats, O. (2023).
\newblock Ride-hailing vs. public transport: Comparing travel time valuation using revealed preference.
\newblock {\em Public Transport: Comparing Travel Time Valuation Using Revealed Preference}.

\bibitem[Yu and Shen, 2019]{yu2019integrated}
Yu, X. and Shen, S. (2019).
\newblock An integrated decomposition and approximate dynamic programming approach for on-demand ride pooling.
\newblock {\em IEEE Transactions on Intelligent Transportation Systems}, 21(9):3811--3820.

\end{thebibliography}

\end{document}